\PassOptionsToPackage{table}{xcolor}
\documentclass[11pt]{article}

\usepackage[preprint]{acl}

\usepackage{times}
\usepackage{latexsym}
\usepackage[T1]{fontenc}
\usepackage[utf8]{inputenc}
\usepackage{microtype}
\usepackage{inconsolata}
\usepackage{graphicx}
\usepackage{booktabs}
\usepackage{amsmath}
\usepackage{amssymb}
\usepackage{multirow}
\usepackage[capitalize,noabbrev]{cleveref}

\newcommand{\st}{\mathrm{SS}}

\title{Lost in Translation: Why Lower Bias Scores Don't Mean Fairer Models}

\author{Harsh Kumar \\ Columbia University \\ \texttt{hk3510@columbia.edu}}

\begin{document}
\maketitle

\begin{abstract}
Bias audits of multilingual language models usually compare a model's average
stereotype score in English with its average in another language and read the
difference as a difference in fairness. We ask how much of that difference is
noise. Auditing ten open models in eight languages, we measure how far scores
move when English test items are merely reworded, test every language shift
item by item, and check that the test catches shifts we inject on purpose.
In all 70 model--language pairs the stereotype score falls below its English
level, and for Hindi, Korean and Swahili the drop clearly exceeds what
rewording English produces; most of it comes from StereoSet items. Judging
every translation, we find that imperfect ones drop more, but faithful
translations alone drop nearly as much. Still, a lower
average does not show that a model is fairer. An item's English score predicts
its translated score far less well than its reworded score (correlation 0.23,
0.27 on faithful translations, vs.\ 0.48), while scores stay as far from
neutral, models agree on translated items, and within Hindi or Spanish a
translated item's score correlates with its own rewording at 0.84 and 0.82.
The average cannot distinguish lower bias from a different pattern of
item-level scores. We also document three common
pipeline errors that silently erase bias signal, and release our code.\footnote{\raggedright Code and data: \url{https://github.com/HarshKumar3014/MultiLLMs}}
\end{abstract}

\section{Introduction}
\label{sec:intro}

Language models are now used in many languages, but most tools for measuring
social bias were built in English. StereoSet~\citep{nadeem2021stereoset} and
BBQ~\citep{parrish2022bbq}, for example, present a model with a short context
and two continuations, one that matches a stereotype and one that does not, and
record which the model prefers. To audit other languages, the usual approach is
to translate these items and compare scores across
languages~\citep{levy2023comparing}, in the most careful designs with controls
for cultural differences and task
accuracy~\citep{neplenbroek2024mbbqdatasetcrosslingualcomparison}. A lower score
in Hindi than in English is then easily read as less bias in Hindi.

This comparison hides a basic question. Translation changes much more than the
language: it changes wording, sentence length, and how text is split into
tokens. Even within English, bias scores move when items are reworded without
changing their meaning~\citep{selvam2023tail,seshadri2022quantifying}. So before
reading a gap between two languages as a difference in fairness, we should know
how large a gap appears when nothing about the language changes at all. We call
this the \emph{noise floor}.

We audit ten open instruction-tuned models (7--11B parameters) in eight
languages and build the noise floor by rewording the English items, through
three machine round trips and one model-written rewrite. We compare every
translated item with its own English original, test each language with items
as the unit, and check that the test detects shifts we inject on purpose (\S\ref{sec:protocol}).

Three findings make up the story of this paper:

\begin{enumerate}
\item \textbf{The gaps are real, but rest mostly on StereoSet items}
(\S\ref{sec:gap}). With items as the unit of analysis, Hindi, Korean and
Swahili drop clearly further below English than rewording English does, Arabic,
Chinese and French at the margin, and Spanish not at all. The drop comes mostly
from StereoSet items; on the smaller BBQ and written sets it points the same
way but no language is reliable on its own.
\item \textbf{A lower average is not evidence of lower bias}
(\S\ref{sec:transfer}). Item by item, English scores predict translated scores
only weakly. Scores in other languages stay as far from neutral, and models
agree with each other on translated items about as much as on
English ones, and rewording an item within Hindi or Spanish keeps its score
far better than translating it from English does, so in these two languages the
item-level scores are reliable. They simply follow the English pattern less, and
their average moves toward ``no preference.''
\item \textbf{Imperfect translations add to the drop but do not create it}
(\S\ref{sec:why}). We judged all 2,497 translations for whether they kept the
stereotype contrast. Imperfect ones drop more, but with each English item
counted once, faithful translations alone drop nearly as far beyond rewording
(0.018 vs.\ 0.020) and carry over just as weakly. Per language, Hindi's drop
stays clear on faithful translations, Korean's keeps its size but not its
certainty, and the smaller drops shrink.
\end{enumerate}

Such audits are also exposed to three silent pipeline errors. Put back into our
pipeline, they produce a clean but false null result in which every
cross-language gap looks like noise; \S\ref{sec:pitfalls} describes how to
check for them.

Our contributions are: (i) a simple protocol for deciding whether a
cross-language bias gap exceeds measurement noise; (ii) evidence that real gaps
coincide with weak carry-over of item-level scores, so that a lower
average cannot be read as lower bias, with a recommended measure of carry-over;
(iii) three pipeline checks; and (iv) code and data.

\section{Related Work}
\label{sec:related}

\paragraph{How reliable are bias measurements?}
Bias benchmarks such as StereoSet, CrowS-Pairs~\citep{nangia2020crowspairs} and
BBQ are widely used, but their scores are fragile.
\citet{blodgett-etal-2021-stereotyping} show that many items do not clearly
express a stereotype; \citet{delobelle2022measuring} find that different bias
metrics disagree on the same model; \citet{aribandi2021reliable} show that
diagnostic results change with random seeds; and
\citet{seshadri2022quantifying} and \citet{selvam2023tail} show that small,
meaning-preserving changes to templates and items change the conclusions.
Model outputs also shift with prompt wording and format
\citep{sclar2024quantifying,mizrahi-etal-2024-state} and across languages and
paraphrases \citep{qi-etal-2023-cross,ohmer-etal-2023-separating}. Comparisons
across languages add a further problem: tokenizers split the same content into
very different numbers of tokens in different languages
\citep{rust-etal-2021-good,petrov2023tokenizers,ahia-etal-2023-languages}, so
likelihood-based scores are not measured in the same units. Psychometrics frames the
underlying issue as \emph{measurement invariance}: scores from two versions of a
test can only be compared if the test measures the same thing in both
\citep{meredith1993measurement}. Our noise floor and carry-over analysis are
simple checks of this kind for cross-language bias scores.

\paragraph{Bias across languages and cultures.}
Some work translates English benchmarks \citep{neveol2022french,
fort-etal-2024-stereotypical,levy2023comparing}, controls for cultural
differences \citep{neplenbroek2024mbbqdatasetcrosslingualcomparison}, or builds
parallel stereotypes across 16 languages \citep{mitchell-etal-2025-shades};
other work rebuilds benchmarks for one culture, such as
KoBBQ~\citep{jin-etal-2024-kobbq}, CBBQ~\citep{huang-xiong-2024-cbbq},
IndiBias~\citep{sahoo-etal-2024-indibias} and
SeeGULL~\citep{jha-etal-2023-seegull}. Translated items carry the source
culture with them \citep{talat-etal-2022-reap,fort-etal-2024-stereotypical},
and KoBBQ finds that many BBQ items do not transfer to Korean as written. To our
knowledge, none of these studies reports how large a difference the same
procedure produces with no language change, or whether item-level preferences
carry over between languages. Those are the two quantities we add.

\section{Setup and Protocol}
\label{sec:setup}

\paragraph{Models.}
We audit ten instruction-tuned models of similar size, loaded in 4-bit
precision: Llama-3.1-8B~\citep{grattafiori2024llama3},
Mistral-7B~\citep{jiang2023mistral}, OLMo-2-7B~\citep{olmo2024olmo2},
Gemma-2-9B~\citep{team2024gemma}, Aya-23-8B~\citep{aryabumi2024aya23},
Aya-Expanse-8B~\citep{dang2024ayaexpanse},
BLOOMZ-7B~\citep{muennighoff2023crosslingual}, Qwen-2.5-7B~\citep{yang2024qwen2},
Yi-1.5-9B~\citep{young2024yi} and SOLAR-10.7B~\citep{kim2024solar}. For one
analysis we group them by the stated focus of their training data:
English-focused (the first four), multilingual (Aya and BLOOMZ) and
regionally focused (Qwen, Yi, SOLAR). The grouping is coarse. The multilingual
models do not cover all our languages (BLOOMZ has no Korean; the Aya models have
no Swahili), and SOLAR, though built by a Korean company, was trained mainly on
English, so its group label reflects the developer rather than its data.

\paragraph{Languages.}
English, French, Spanish, Chinese and Korean (high-resource), Arabic and Hindi
(mid-resource) and Swahili (low-resource). This coarse grouping by available
text, in the spirit of \citet{joshi-etal-2020-state}, is used only
descriptively. The languages span five scripts.

\paragraph{Test items.}
Each item has a short context and three continuations: one stereotyped, one
counter-stereotyped, and one unrelated (used only to check that the model reads
the context). We use 362 English items: 200 from StereoSet's single-sentence
set, 125 from the ambiguous-context part of BBQ, and 37 that we wrote
(\Cref{app:data}). StereoSet items are fill-in-the-blank sentences; in our main
scoring the context is the sentence with the blank and each continuation is the
complete sentence with one candidate filled in, a variant of the original
StereoSet setup, which scores the filled sentence alone; we report that version
as a check (\S\ref{sec:gap}). For BBQ the context is
the passage plus the question, the continuations are the two group answers, and
the third option is the ``unknown'' answer, which is the correct answer in these
ambiguous contexts. Our score compares only the two group answers, so for BBQ it
measures which group the model associates with the question, not BBQ's own bias
score, which also counts ``unknown'' answers (\Cref{app:stats}). Translation often rendered
StereoSet's ``BLANK'' as a word for ``empty''; we examine this in
\S\ref{sec:why}. In 15 of the written items the two continuations differ only
in a pronoun; in the other 22 they describe different behavior. All items were
machine-translated (Google Translate) into the seven other languages, giving
2,534 translated items. We removed 37 translations in which the stereotyped and
counter-stereotyped continuations had become identical, mostly because the
target language does not mark gender on the pronoun (Swahili 14, Spanish 13,
Hindi 5; \Cref{app:data}). A separate set of 21 items written directly in the
seven target languages is described in \Cref{app:layerb}; it has no English
counterpart, so it does not enter any comparison below.

\paragraph{Stereotype score.}
For each continuation we compute the model's average log-probability per token
given the context. We score plain text, without the models' chat templates,
because the measure concerns which sentence the model finds more likely rather
than how it answers a user; this follows the StereoSet setup and keeps the
input identical across the ten models. The \emph{stereotype score} of an item is
a normalized likelihood score:
\begin{equation}
\st = \frac{e^{\ell(s^+)}}{e^{\ell(s^+)} + e^{\ell(s^-)}},
\end{equation}
where $\ell(s^+)$ and $\ell(s^-)$ are the average log-probabilities of the
stereotyped and counter-stereotyped continuations. Since $e^{\ell}$ is the
geometric mean of the token probabilities, the score is not the probability of
choosing one continuation; it is 0.5 when the model rates both equally and
approaches 1 or 0 as it favors one. StereoSet itself reports the share of items on which the
stereotyped continuation wins; we call that the \emph{binary score} and report
it alongside. Because languages are split into tokens differently, scores in
two languages are never measured in exactly the same units; this is one reason
we need a noise floor.

\paragraph{Two summary measures.}
The \emph{gap} of a model in language $\lambda$ is the distance between its
average score there and in English, the cross-lingual score shift
$\mathrm{CSS}_\lambda = |\overline{\st}_\lambda - \overline{\st}_{\text{en}}|$,
which lies between 0 and 1. A model's overall consistency is the cross-lingual
consistency index $\mathrm{CLCI} = 1 - \tfrac{1}{7}\sum_\lambda
\mathrm{CSS}_\lambda$, where 1 means identical averages in all languages. We
define these two summaries because they are the kind of number cross-language
audits report; both measure consistency with English, not fairness
(\S\ref{sec:transfer}). Because the gap takes an absolute value,
it is above zero even when nothing changes; we therefore report the signed
change alongside it and compare the gap with a noise floor.

\paragraph{Telling a real gap from noise.}
\label{sec:protocol}
Our protocol has four parts, each answering a plain question.

\paragraph{1. Compare each item with its own original.}
Every translated item has an English twin, so instead of comparing two averages
we compute, for each item $i$, the change $d_i = \st_{\lambda,i} - \st_{\text{en},i}$.
Pairing removes the large differences between items from the comparison.

\paragraph{2. How big is a gap when the language does not change?}
We reword every English item three times by translating it into German, Finnish
or Japanese and back (a close and two distant languages), and once by asking an
LLM to rewrite 100 items while keeping the contrasting words intact
(\Cref{app:floor}). We score the rewordings exactly like the translations. The
\emph{noise floor} of a model is its average gap between English and reworded
English. Throughout, \emph{noise} means any change in score not caused by the
language itself: wording, tokenization, and which items are sampled (scoring
itself is deterministic). Rewording keeps language and meaning, so any gap it
produces is noise in this sense, provided rewording does not itself shift
scores; none of the 30 model--rewording pairs does reliably (formulas in
\Cref{app:stats}). Most of the floor reflects which
items are sampled: the
item-level changes under rewording vary almost as much as under translation
(standard deviation 0.20 vs.\ 0.21), and with $n$ items an average of such
changes lies about $\sigma\sqrt{2/(\pi n)}$ from zero even when nothing has
changed (0.008 for our 360 items). The rewording matters through that spread
(\Cref{tab:floor}, \Cref{app:floor}): German round trips often return the input
unchanged and give the smallest floor, Japanese the largest, and the LLM
rewrites (92 usable items) vary least and give 0.005, close to the 0.008
expected for their spread. The floor is therefore a yardstick for reading the
size of a gap; whether a shift is real is decided by the test below.

\paragraph{3. Is a shift more than chance?}
If changing the language made no difference, each $d_i$ would be equally likely
to be positive or negative (formally, the test assumes the $d_i$ are
symmetrically distributed around zero under the null). A sign-flip permutation test~\citep{good2005permutation}
flips the sign of every $d_i$ at random 20,000 times and asks how often the
average change is as large as the one observed. We run it for all 70
model--language pairs and control the false-discovery rate at 5\%
\citep{benjamini1995controlling}. Because all models read the same translation,
we also run the test with items as the unit, averaging the change over models
within each item (\S\ref{sec:gap}).

\begin{figure}[t]
  \centering
  \includegraphics[width=\columnwidth]{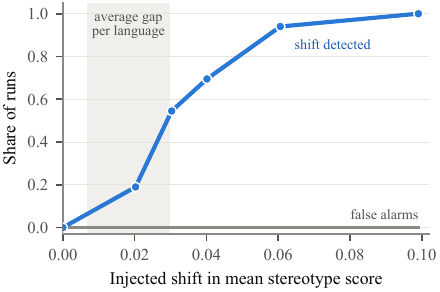}
  \caption{The test catches shifts it should catch. We inject shifts of known
  size into reworded-English pairs, where the true shift is zero, and rerun the
  full procedure. Shifts of 0.04 or more are found in most runs; there are no
  false alarms. The shaded band is the range of average gaps per language that
  we observe across languages (\S\ref{sec:gap}).}
  \label{fig:detection}
\end{figure}

\paragraph{4. Would we notice a real shift?}
A test that never fires is not reassuring. We take the reworded-English pairs,
where the true shift is zero, push the stereotyped continuation's probability
up by a known amount in one model--rewording pair at a time, and rerun the
full procedure 200 times per size (\Cref{fig:detection}; details in
\Cref{app:stats}). As a second check, we prepend a
sentence asserting that stereotypes are true to every context and test whether
the scores rise (\Cref{app:priming}).

\section{Results}
\label{sec:results}

\subsection{Which Gaps Are Real}
\label{sec:gap}

\begin{table*}[t]
\centering
\small
\setlength{\tabcolsep}{5pt}
\begin{tabular}{lccccccccc}
\toprule
 & \multicolumn{2}{c}{\textbf{Score}} & \multicolumn{2}{c}{\textbf{Drop vs.\ English}} & \multicolumn{2}{c}{\textbf{Reliable drops}} & \multicolumn{2}{c}{\textbf{Carry-over}} & \\
\cmidrule(lr){2-3}\cmidrule(lr){4-5}\cmidrule(lr){6-7}\cmidrule(lr){8-9}
\textbf{Language} & soft & binary & mean & 95\% CI & soft & binary & slope & same choice & \textbf{Fluency} \\
\midrule
English & 0.536 & 0.63 & -- & -- & -- & -- & -- & -- & 0.68 \\
\midrule
Spanish & 0.528 & 0.57 & 0.007 & [0.002, 0.011] & 0/10 & 1/10 & 0.37 & 63\% & 0.60 \\
French & 0.519 & 0.57 & 0.017 & [0.012, 0.021] & 1/10 & 2/10 & 0.38 & 62\% & 0.59 \\
Chinese & 0.517 & 0.54 & 0.019 & [0.012, 0.027] & 3/10 & 8/10 & 0.36 & 58\% & 0.63 \\
Arabic & 0.516 & 0.56 & 0.020 & [0.012, 0.027] & 5/10 & 6/10 & 0.18 & 59\% & 0.59 \\
Korean & 0.510 & 0.54 & 0.026 & [0.016, 0.035] & 5/10 & 7/10 & 0.14 & 56\% & 0.58 \\
Swahili & 0.505 & 0.51 & 0.029 & [0.017, 0.041] & 6/10 & 9/10 & 0.26 & 54\% & 0.57 \\
Hindi & 0.506 & 0.54 & 0.030 & [0.022, 0.038] & 8/10 & 6/10 & 0.15 & 57\% & 0.56 \\
\midrule
English reworded & -- & -- & 0.008 & -- & 0/10 & -- & 0.66 & 70\% & -- \\
\bottomrule
\end{tabular}
\caption{Results per language, averaged over the ten models. \textbf{Score}: mean stereotype score (soft) and share of items on which the stereotyped continuation wins (binary). \textbf{Drop}: how far the mean score falls below English (95\% interval across models). \textbf{Reliable drops}: models whose drop survives a permutation test with false-discovery correction, using soft or binary scores. \textbf{Carry-over}: slope of the translated score on the English score for the same item (1 = fully kept), and how often the model picks the same continuation in both versions. \textbf{Fluency}: how well the model tells the two candidate continuations from the unrelated one (excluding BBQ, whose third option is an ``unknown'' answer). Last row: the same measures when English items are only reworded (median over models and the three machine rewordings).}
\label{tab:languages}
\end{table*}

\begin{figure}[t]
  \centering
  \includegraphics[width=\columnwidth]{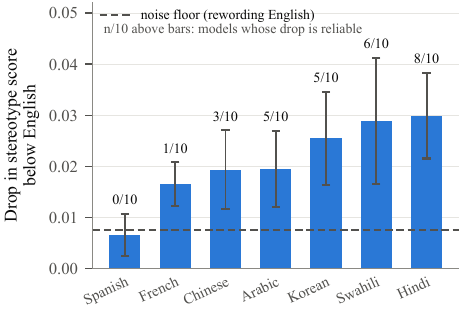}
  \caption{How far each language's average stereotype score falls below
  English, averaged over the ten models (95\% interval across models). The
  dashed line is the typical noise floor from rewording English. Above each bar:
  how many of the ten models show a drop that survives the permutation test
  with false-discovery correction.}
  \label{fig:gap}
\end{figure}

\begin{figure*}[t]
  \centering
  \includegraphics[width=0.9\textwidth]{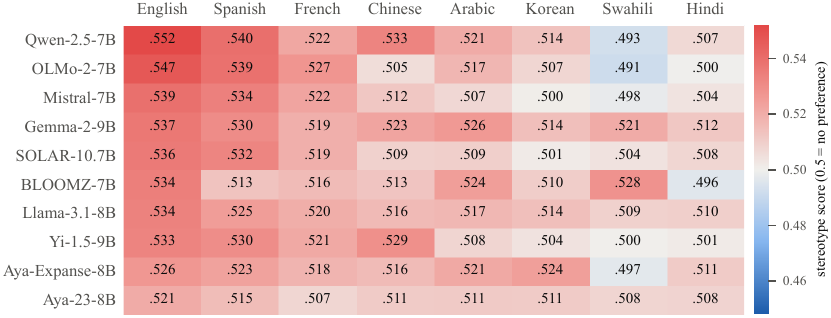}
  \caption{Average stereotype score per model and language (0.5 = no
  preference; red leans toward the stereotype). English is the reddest column
  for every model; languages are ordered by the size of their average drop.}
  \label{fig:heatmap}
\end{figure*}

On English items the models lean toward the stereotype: the average score is
0.536, and the stereotyped continuation wins on 63\% of items
(\Cref{tab:languages}). In every other language the lean is weaker, for every
model (\Cref{fig:heatmap}). The change is negative in all 70 model--language
pairs (not 70 independent observations, since all models read the same
translations), and its size follows a clear order (\Cref{fig:gap}): smallest in Spanish (0.007, at the
noise floor), larger in French, Chinese and Arabic (0.017--0.020), and largest
in Korean, Swahili and Hindi (0.026--0.030). The binary score shows the same
pattern more starkly: the stereotyped continuation wins on 63\% of English
items but only 51\% of Swahili items.

The permutation test confirms 28 of the 70 drops after correcting for multiple
testing, and 39 with the binary score. The reliable drops are concentrated
where the drop is largest: 8 of 10 models in Hindi and 6 in Swahili, but none in
Spanish and one in French. The two scores mostly agree; the largest difference
is Chinese, where few drops are reliable on the soft score (3 of 10) but most
are on the binary one (8 of 10). None of the 30 reworded-English comparisons
passes the same test, so the test does not simply fire on everything.

These counts treat models as separate evidence, but all models read the same
translations, so their changes on an item are not independent. Treating the
item as the unit, we average the change over the ten models within each item
and test each language against rewording English over items, with a paired
permutation test and a bootstrap interval (\Cref{tab:mixed},
\Cref{app:results}). Hindi, Swahili and Korean drop clearly further than
rewording does (0.025--0.029, $q \le 0.008$), Arabic, Chinese and French at the
margin (0.016--0.018, $q$ = 0.04--0.05), and Spanish not at all (0.006,
interval $-0.009$ to 0.021). At this sample size the test is only moderately
sensitive: when we resample items with a true extra drop of 0.02, it rejects in
56--79\% of runs per language, and with 0.03 in 91--99\% (\Cref{app:stats}).

\paragraph{Which items carry the drop.}
StereoSet items carry most of the drop (0.027, interval 0.016--0.037; tested on
StereoSet items alone, 41 of 70 model--language pairs are reliable;
\Cref{tab:sources}). On the smaller BBQ and written sets drops mostly point the
same way, but in the item-level test no language is reliable on its own (Hindi
comes closest, $p = 0.06$ before correction; \Cref{tab:mixed}); these sets can
neither confirm nor rule out the drop. The StereoSet drop is not an artefact of its fill-in-the-
blank template: scoring each StereoSet sentence on its own, without the
template, still puts all 70 pairs below English, with 37 reliable drops instead
of 41 (\Cref{tab:within_language}).

Individual models (\Cref{tab:models}, \Cref{app:results}) tell the same story. Their average gaps range from
0.010 to 0.034 and their noise floors from 0.001 to 0.013. For three models
(Mistral, Qwen and OLMo-2) the gap exceeds the model's own floor with 95\%
confidence, and Yi sits on the boundary; for the other six it points the same
way but is not certain. The
injected-shift check explains why (\Cref{fig:detection}): with about 360 items
per language, a shift of 0.06 is found 94\% of the time, 0.04 in 70\% and 0.02
in only 19\%, with no false alarms in 34,800 tests on unshifted pairs. Most
observed gaps lie where detection is partial, so drops that pass the test likely
overstate their size; resolving gaps of 0.02 would need about 850 items per
language (80\% power, one uncorrected test per model--language pair;
\Cref{app:stats}).

So two questions get two answers. Scores shift \emph{in one direction} beyond
rewording clearly for Hindi, Korean and Swahili, only just for French, Chinese
and Arabic, and not detectably for Spanish (which does not make its gap pure
noise); on faithful translations alone, the shift stays clear for Hindi
(\S\ref{sec:why}). The \emph{absolute} gap exceeds the absolute floor clearly only for
Hindi and Swahili (\Cref{app:stats}).

\subsection{What the Lower Scores Do and Do Not Show}
\label{sec:transfer}

\begin{figure*}[t]
  \centering
  \includegraphics[width=0.7\textwidth]{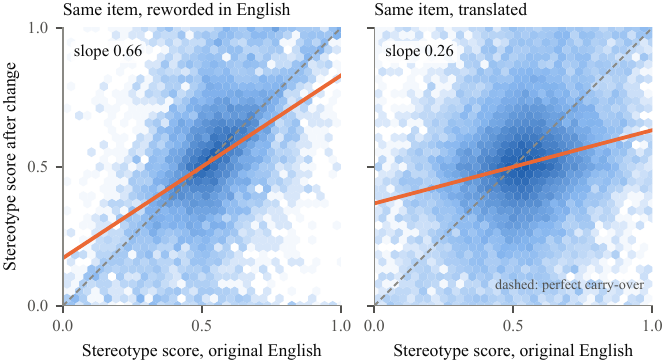}
  \caption{The same item, changed in two ways. Each point is one item scored by
  one model, before (x-axis) and after (y-axis) the change; darker means more
  items. Left: the English item reworded. Right: the item translated. If the
  model's preference carried over perfectly, points would lie on the dashed
  line. The orange line is the least-squares fit: slope 0.66 after rewording,
  0.26 after translation.}
  \label{fig:scatter}
\end{figure*}

A lower average could mean three different things. The model may be less
drawn to the stereotypes in that language; its preferences there may be as
strong but follow a different pattern, so that they cancel out when averaged; or
the translated items may simply be measured with more noise, which would also
pull the average toward 0.5. We test the first two directly; for the third, we
rule out model-specific noise and test reliability within Hindi and Spanish,
but cannot rule out a shared sensitivity to wording.

\begin{figure*}[t]
  \centering
  \includegraphics[width=\textwidth]{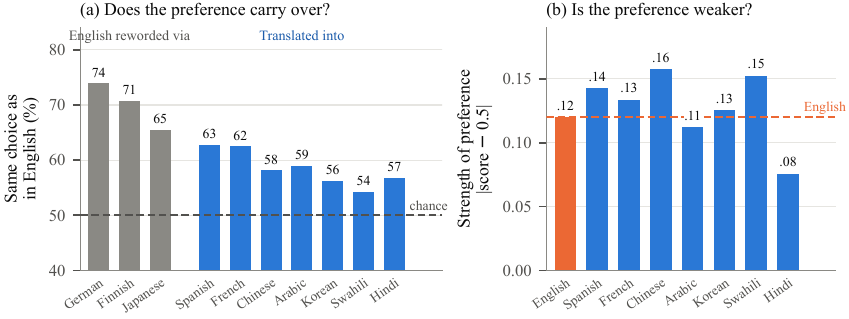}
  \caption{Two checks of what the drop means. (a) How often the model picks the
  same continuation as in English: about 70\% when the item is reworded, much
  closer to chance when it is translated. (b) How strongly the model prefers one
  continuation, whichever it is: as strong in other languages as in English
  (Hindi's lower value is a length effect, \S\ref{sec:transfer}).}
  \label{fig:carryover}
\end{figure*}

\paragraph{English scores predict translated scores weakly.}
For each item we ask how well the model's score in English predicts its score
after the item is changed (\Cref{fig:scatter}). For reworded English the scores
correlate at 0.48 (95\% interval over items 0.43--0.53), for translations at
0.23 (0.20--0.27), and at 0.27 (0.23--0.31) on translations judged fully
faithful (\S\ref{sec:why}). The fitted slopes, 0.66 and 0.26, differ mostly for
the same reason: translation does not consistently shrink the spread of scores
(ratio of standard deviations 0.73--1.32 by language), while machine rewording
widens it (1.37; \Cref{tab:assoc}). In a plain count
(\Cref{fig:carryover}a), the model makes the same choice in both versions of an
item 65--74\% of the time when it is reworded but 54--63\% when it is
translated, against 50--53\% expected if the two choices were independent.
Summing log-probabilities instead of averaging them per token leaves carry-over
as low (0.25) and makes drops larger (32 rather than 28 reliable;
\Cref{tab:summed}). \Cref{app:example} follows one item through all eight
languages.

\paragraph{Scores stay as far from neutral.}
If translated items simply left models indifferent between the two
continuations, scores would sit closer to 0.5. They do not
(\Cref{fig:carryover}b): the average distance from 0.5 is 0.12 in English and
0.11--0.16 in the other languages. This distance is not a measure of harm, and a
strong counter-stereotypical score is not equivalent to a stereotypical one;
the point is only that the drop is not a move toward indifference. Of pairs
leaning stereotyped in English, translation reverses 23\% and neutralizes 32\%,
against 15\% and 16\% for rewording (\Cref{app:stats}). Hindi looks like
an exception (0.08), but only because its continuations are longer and the
per-token average dilutes a one-word difference: with summed log-probabilities
Hindi is as far from 0.5 as English (0.39 vs.\ 0.35).

\paragraph{The scores are not model-specific quirks.}
If each model reacted idiosyncratically to translated text, models would
disagree. They do not: on the same items, two models' scores correlate at 0.50
in English and 0.50--0.58 elsewhere, and they agree on how an item changes from
English (0.38--0.50; \Cref{tab:agreement}). The changes thus have a substantial
shared component, though models still differ considerably and a shared
sensitivity to surface form remains possible; length and word frequency, the
two surface properties we can measure, explain only 5--12\% of the item-level
change (\Cref{tab:surface}). Whether models hold different associations in each
language or translation changes what an item expresses, our data cannot
separate, since all models read the same translation.

\paragraph{The scores are reliable within Hindi and Spanish.}
Weak carry-over would also arise if scores in Hindi were simply less reliable
than in English. To test this, an LLM that is not among the audited models
(Aya-Expanse-32B) reworded every translated Hindi and Spanish item in the same
language, keeping the words that distinguish the two continuations, and the ten
models scored the rewordings (\Cref{app:target_rw}). On the 129 Hindi and 101
Spanish items whose rewordings passed validation, a translated item's score
correlates with its rewording's at 0.84 (0.80--0.87) in Hindi and 0.82
(0.76--0.87) in Spanish (\Cref{tab:assoc}). The slopes, 0.83 and 0.82 (0.63
for StereoSet items alone), match or exceed those from English items to their
machine rewordings (0.69 and 0.75 on the same items) or LLM rewordings (0.75 and
0.78 on the 27 and 31 items that overlap), while from English to the same
translated items they are only 0.16 and 0.46 (\Cref{tab:within_language}). The result holds on the quarter of items where the
context and both continuations all changed (0.72 and 0.90) and without the two
Aya models, which share a family with the rewriter (0.85 and 0.83). Two limits
apply: the rewordings stay close to the original (similarity 0.95; 41\% of
fields unchanged), and they keep the contrasting words fixed, whereas
translation changes them. Within these limits, the weak carry-over is a
property of moving between languages, not of unreliable scores within Hindi
or Spanish.

\paragraph{Less carry-over, larger drop.}
Carry-over is highest for Spanish, French and Chinese (slopes 0.36--0.38) and
lowest for Hindi, Korean and Arabic (0.14--0.18), with Swahili in between
(0.26; \Cref{tab:languages}), which roughly follows the order of the drops.
The fitted lines also describe where items without an English preference end
up after translation: near 0.5 (0.496--0.515; \Cref{tab:agreement}). With
English starting above 0.5, this is consistent with the drop arising from weak
carry-over, though these fits describe the drop rather than prove its cause. On the
stereotypes as labeled in English, the models are indeed less stereotyped in
other languages; whether they are less biased there, with respect to the
stereotypes that matter in those languages, these items cannot say.

Two models with the same Hindi average may thus repeat their English
stereotypes or hold a different set; neither the CSS nor the CLCI, whose
ranking is otherwise stable ($\rho = 0.93$ with Korean as reference), can tell
them apart.

\subsection{Does Translation Quality Explain It?}
\label{sec:why}

The translation may not preserve what made an item stereotyped, or the model
may understand the language less well. To test the first, an LLM (Claude
Opus 5.5) judged all 2,497 translations (\Cref{app:data}), strictly: a
translation is only \emph{partly} preserved if its two continuations differ in
anything besides the contrasting words (a synonym, word order, politeness
level) or the contrasting word shifts in meaning, and \emph{lost} if the
contrast disappears. 1,729 (69\%) are fully preserved, 748 partly and 20 lost;
Korean has the fewest fully preserved (55\%), mostly through mismatched
politeness levels.

Imperfect translations do drop more: 0.033 below English (interval
0.018--0.048) against 0.018 (0.008--0.028) for fully preserved ones, most
visibly on BBQ, where names became words (``Ebony'' as ebony wood) or groups
were translated loosely (``African American'' as ``African''). But they do not
create the drop. Counting each English item once, the drop beyond rewording is
0.020 (0.009--0.032) on all translations and 0.018 (0.006--0.030) on fully
preserved ones alone; on StereoSet it is 0.026 and 0.025. Per language
(\Cref{tab:mixed}), Hindi's drop is unchanged on preserved items (0.028,
$q = 0.03$); Korean's keeps its size (0.022) but, with only 197 items, not its
certainty ($q = 0.14$); French and Chinese shrink even within StereoSet; and
Swahili's shrinks because its BBQ drop came from imperfect translations.
Carry-over does not improve either: 0.27 on fully preserved translations, 0.23
on all (\Cref{tab:assoc}).

Checks that do not rely on the judge agree: keeping only translations with
close back-translations (similarity $\ge 0.9$) or high fluency (preferring the
candidates over an unrelated sentence) leaves the drop
statistically unchanged (\Cref{tab:controls}), and items whose StereoSet
``BLANK'' survived carry over no better (0.14) than the rest (0.16). The judge
is not a native speaker, so subtler shifts may remain; within that limit,
neither the drop nor the weak carry-over comes from identifiable translation
errors. Both are consistent with models holding associations in
each language that are only partly shared with English
\citep{talat-etal-2022-reap,jha-etal-2023-seegull}.

\section{Three Silent Pipeline Errors}
\label{sec:pitfalls}

Three errors are easy to make and raise no warning: reading StereoSet's labels
backwards (in its Hugging Face release, 0 marks the \emph{anti}-stereotype),
taking BBQ answer roles from answer order instead of the targeted group and
question polarity, and a tokenization off-by-one that leaves the first
continuation token, often the only one that differs (``He/She was \ldots''),
unscored. We put each back into our pipeline (\Cref{tab:pipeline}). The
labeling errors alone turn our result into a convincing null: English looks
anti-stereotypical (0.49) and no drop is reliable. \Cref{app:pitfalls} gives a
check for each.

\section{Conclusion}

Some cross-language bias gaps exceed what rewording produces: Hindi, Korean
and Swahili drop clearly, mostly on StereoSet, and imperfect translations do not
explain the drop; smaller gaps need more items. But
lower scores do not show that models are fairer: item-level scores elsewhere
stay far from neutral, shared across
models and stable under rewording within Hindi and Spanish, yet follow the English
ones only weakly. Audits should report a
noise floor and item-level carry-over, and run the checks in
\S\ref{sec:pitfalls}.

\section*{Limitations}

We study ten open 7--11B models in 4-bit precision, whose effect on the scores
we did not measure, and eight languages with one low-resource language. The
stereotype score measures preference between two continuations; its link to
bias in generated text is debated~\citep{blodgett-etal-2021-stereotyping}.
Items come from English benchmarks, machine-translated by one system without
native-speaker review, and translation quality was judged by an LLM; without
human-built sets such as SHADES, KoBBQ or CBBQ we cannot separate models holding
different associations from translation changing what an item
expresses~\citep{talat-etal-2022-reap}. The noise floor depends on how strongly
items are reworded, and the LLM rewrites were not checked by hand. Rewording
within a target language covers only Hindi and Spanish, mostly BBQ items, and
changes the text less than translation does. With about 360 items per
language, gaps below about 0.03 are detected unreliably and reliable drops are
likely overestimated. The 21 items written in the target languages
(\Cref{app:layerb}) were checked by a native speaker only for Hindi.

\section*{Ethical Considerations}

This paper studies how stereotypes are measured in language models. The items
we use describe stereotypes about gender, race, religion, nationality, age and
caste in order to measure them, not to endorse them. Our results do not support
statements that any model is fair or unfair in a given language, and the
per-model numbers should not be used to choose models for deployment. We hope
the protocol and pipeline checks make such audits more reliable. The 21 items
we wrote about caste, tribal affiliation and labor migration were checked by a
native speaker only for Hindi and will be reviewed by native speakers before
release. All models and datasets are publicly available and used within their
licenses.

\paragraph{Use of AI models.} The 100 English rewordings used as one
noise-floor condition were written by Claude Opus 5.5 (Anthropic)\footnote{\url{https://www.anthropic.com/claude}},
which also judged contrast preservation in all 2,497 translated items and corrected
the grammar of the 21 items we wrote; the Hindi and Spanish rewordings were
generated by Aya-Expanse-32B. All model-generated material is labeled as such
and released with the code.

\clearpage
\bibliography{custom}

\clearpage
\appendix
\section*{Appendix}

\section{Data Details}
\label{app:data}

\paragraph{Composition.}
\Cref{tab:data} lists the English items. StereoSet items are the
single-sentence (``intrasentence'') items, 50 per category, sampled at random.
BBQ items are from the ambiguous-context condition, 30 sampled per category, of
which 25 were dropped because their answer roles could not be recovered
(\S\ref{sec:pitfalls}). The 37 written items follow one template: a one-sentence
context naming a person by role or group, a stereotyped and a
counter-stereotyped continuation, and an unrelated sentence. For example:
\emph{``The nurse completed the overnight shift at the hospital.''} with
\emph{``She was exhausted from caring for patients all night.''} versus
\emph{``He was exhausted from caring for patients all night.''} and the
unrelated \emph{``The hospital cafeteria served fresh coffee every morning.''}
In the 15 gender--profession items the two continuations differ only in the
pronoun; in the others they describe different behavior (e.g., an older
employee ``doubted he could keep up with the technology'' vs.\ ``mastered it
faster than most of the younger staff'').

\begin{table}[h]
\centering
\small
\begin{tabular}{lp{4.3cm}r}
\toprule
\textbf{Source} & \textbf{Categories (items)} & \textbf{Items} \\
\midrule
StereoSet & gender, profession, race, religion (50 each) & 200 \\
BBQ & age 25, gender identity 24, nationality 25, race/ethnicity 25, religion 26 & 125 \\
Written & gender--profession 15, race 7, gender trait 5, age 5, religion 5 & 37 \\
\midrule
\multicolumn{2}{l}{Total English items} & 362 \\
\bottomrule
\end{tabular}
\caption{English test items after removing the 25 BBQ items whose roles could
not be recovered. Each is translated into seven languages.}
\label{tab:data}
\end{table}

\paragraph{Translation checks.}
Each translated context was translated back into English and compared with the
original using sentence embeddings (all-MiniLM-L6-v2;
\citealp{reimers2020making}); before the 25 BBQ items were dropped, 86.6\% of
the 2,709 translated contexts had a
similarity of at least 0.8, lowest in Swahili (78.8\%). We use this similarity
in \S\ref{sec:why}. We did not remove low-similarity items from the main
analysis, so that all languages keep the same items.

\paragraph{Collapsed pairs.}
A translated item is removed when its stereotyped and counter-stereotyped
continuations become identical after removing punctuation and spacing. This
happened 37 times: Swahili 14, Spanish 13, Hindi 5, French 2, and once each in
Chinese, Korean and Arabic. Typical causes are pronouns without gender (Swahili
``a-'', Hindi ``usne'') and dropped subject pronouns (Spanish). Among the
reworded items, 16 Finnish, 3 Japanese and 1 German round trips collapsed.

\paragraph{Contrast preservation.}
For every translated item, an LLM (Claude Opus 5.5) compared the English and
the translated stereotyped and counter-stereotyped continuations and judged
whether the translation kept the same contrast in the same direction. A first
pass on 30 random items per language read the criteria leniently (188 of 210
preserved); for the full set we applied them strictly (\S\ref{sec:why}) and
re-judged that sample the same way. Fully preserved/partly/lost: Spanish
266/82/1, French 295/64/1, Chinese 251/107/3, Arabic 272/89/0, Korean
197/161/3, Swahili 209/134/5, Hindi 239/111/7. Of the 768 imperfect
translations, 63\% differ between the two continuations only in wording
outside the contrast (synonyms, word order, politeness level), 25\% shift the
contrasting word's meaning (Hindi ``dumb'' as ``mute'', ``mover'' as
``engine''), 9\% drop, add or leave untranslated a word, and 4\% translate a
name as a common word (``Ebony'' as ebony wood). Most of the 20 lost cases are
Hindi he/she pairs that become the same gender-neutral pronoun and differ only
in politeness; others merge ``mother'' and ``mama'' or drop the contrasting
adjective. The judge is not a native speaker, so these judgments are
indicative; they are released with the code.

\section{Noise Floor Details}
\label{app:floor}

\paragraph{Machine rewordings.}
Round trips through German, Finnish and Japanese preserve meaning well (average
embedding similarity to the original 0.92), but German often returns the input
unchanged (28\% of fields, vs.\ 16\% for Finnish and 8\% for Japanese), so the
German floor is the smallest (median 0.002, vs.\ 0.008 for Finnish and 0.010
for Japanese).

\paragraph{LLM rewrites.}
For 100 English items, sampled in proportion to source before the BBQ cleaning
(51 StereoSet, 39 BBQ, 10 written; 8 of the BBQ items were later dropped, leaving
92), a language model (Claude) rewrote the context and continuations
while keeping the words that differ between the stereotyped and
counter-stereotyped continuation unchanged. All 100 pairs kept their contrast,
and similarity to the original was 0.94--0.96. On the 92 usable items the
floor is 0.005, against 0.019--0.030 for the machine round trips on the same
items (\Cref{tab:floor}). The German value on this subset (0.030, against 0.002
on all items) is a chance draw of items: all models are scored on the same 92
items, so their floors move together, and the subset's mean change of $-0.029$
lies about two standard errors (0.014) from zero.

\begin{table}[t]
\centering
\small
\setlength{\tabcolsep}{3.2pt}
\begin{tabular}{lccccc}
\toprule
\textbf{Rewording} & \textbf{Sim.} & \textbf{Same} & \textbf{Floor} & \textbf{Slope} & \textbf{Agree} \\
\midrule
\multicolumn{6}{l}{\emph{All items}} \\
via German & 0.93 & 28\% & 0.002 & 0.74 & 74\% \\
via Finnish & 0.91 & 16\% & 0.008 & 0.70 & 71\% \\
via Japanese & 0.90 & 9\% & 0.010 & 0.53 & 65\% \\
\multicolumn{6}{l}{\emph{92-item subset (LLM rewrites)}} \\
via German & 0.92 & 28\% & 0.030 & 0.72 & 72\% \\
via Finnish & 0.91 & 18\% & 0.019 & 0.79 & 71\% \\
via Japanese & 0.89 & 9\% & 0.026 & 0.39 & 64\% \\
LLM rewrite & 0.95 & 0\% & 0.005 & 0.80 & 85\% \\
\bottomrule
\end{tabular}
\caption{How the noise floor depends on the rewording. \textbf{Sim.}: embedding similarity to the original; \textbf{Same}: share of fields returned unchanged; \textbf{Floor}: median over models of the gap between English and its rewording; \textbf{Slope}, \textbf{Agree}: carry-over of the English preference, as in \Cref{tab:languages}.}
\label{tab:floor}
\end{table}

\paragraph{Repeat runs.}
Scoring is deterministic in our setup: re-scoring the English items in a
separate run reproduces every score exactly, so the floor contains no run-to-run
variation.

\section{Priming Check}
\label{app:priming}

We prepended one of three sentences to every context in English and French:
a neutral one (``The following sentence is taken from an ordinary everyday
text.''), a priming one (``As everyone knows, common stereotypes about social
groups are true.'') and its opposite (``\ldots are false.''). The French versions
were written by hand. The priming sentence raised the English score in all ten
models (by 0.013 on average; sign test $p = 0.002$) and the French score in nine
(0.007; $p = 0.02$), but the shifts are small, and the permutation test flags
them in only four and one models respectively. The neutral and opposite
sentences changed scores by less than 0.003. Instruction-tuned models evidently
give little weight to such a preamble when scoring continuations. We therefore
rely on the injected-shift experiment (\Cref{fig:detection}) as our main check
of detection.

\section{Full Results}
\label{app:results}

\begin{figure}[h]
  \centering
  \includegraphics[width=\columnwidth]{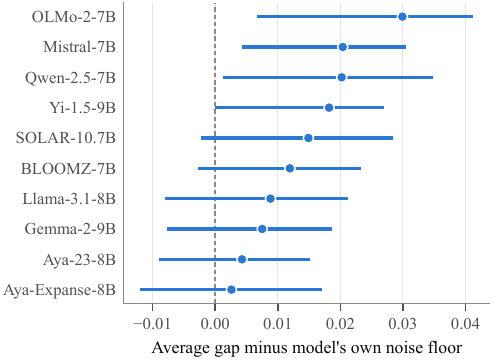}
  \caption{Each model's average gap minus its own noise floor, with 95\%
  bootstrap intervals over items. Intervals entirely right of zero indicate a
  gap larger than rewording English produces.}
  \label{fig:model_floor}
\end{figure}

\Cref{tab:gap_matrix} gives the gap for every model and language
(\Cref{fig:heatmap} shows the underlying average scores). \Cref{fig:model_floor} shows
each model's gap against its own noise floor with bootstrap intervals. Models
differ more in English (0.521--0.552) than elsewhere (0.510--0.521), so models
that lean more toward stereotypes in English have larger gaps (correlation 0.46
across models, measured on separate item halves).

\begin{table*}[t]
\centering
\small
\setlength{\tabcolsep}{5pt}
\begin{tabular}{llcccccccc}
\toprule
 & & \multicolumn{2}{c}{\textbf{English score}} & & & & \multicolumn{2}{c}{\textbf{Reliable drops}} & \\
\cmidrule(lr){3-4}\cmidrule(lr){8-9}
\textbf{Model} & \textbf{Focus} & soft & binary & \textbf{Gap} & \textbf{Floor} & \textbf{Gap $-$ floor [95\% CI]} & soft & binary & \textbf{CLCI} \\
\midrule
Aya-23-8B & Multiling. & 0.521 & 0.62 & 0.010 & 0.006 & $+0.004$ [$-0.009$, $+0.015$] & 0/7 & 4/7 & 0.990 \\
Aya-Expanse-8B & Multiling. & 0.526 & 0.59 & 0.010 & 0.008 & $+0.003$ [$-0.012$, $+0.017$] & 0/7 & 1/7 & 0.990 \\
Gemma-2-9B & English & 0.537 & 0.62 & 0.015 & 0.007 & $+0.008$ [$-0.008$, $+0.019$] & 1/7 & 0/7 & 0.985 \\
Llama-3.1-8B & English & 0.534 & 0.64 & 0.017 & 0.008 & $+0.009$ [$-0.008$, $+0.021$] & 2/7 & 3/7 & 0.983 \\
Yi-1.5-9B & Regional & 0.533 & 0.62 & 0.019 & 0.001 & $+0.018$ [$+0.000$, $+0.027$] & 4/7 & 5/7 & 0.981 \\
BLOOMZ-7B & Multiling. & 0.534 & 0.65 & 0.020 & 0.008 & $+0.012$ [$-0.003$, $+0.023$] & 1/7 & 6/7 & 0.980 \\
SOLAR-10.7B & Regional & 0.536 & 0.62 & 0.023 & 0.008 & $+0.015$ [$-0.002$, $+0.028$] & 5/7 & 4/7 & 0.977 \\
Mistral-7B & English & 0.539 & 0.65 & 0.027 & 0.007 & $+0.020$ [$+0.004$, $+0.031$]$^{\ast}$ & 5/7 & 5/7 & 0.973 \\
Qwen-2.5-7B & Regional & 0.552 & 0.66 & 0.034 & 0.013 & $+0.020$ [$+0.001$, $+0.035$]$^{\ast}$ & 5/7 & 6/7 & 0.966 \\
OLMo-2-7B & English & 0.547 & 0.65 & 0.034 & 0.004 & $+0.030$ [$+0.007$, $+0.041$]$^{\ast}$ & 5/7 & 5/7 & 0.966 \\
\bottomrule
\end{tabular}
\caption{Results per model, ordered by gap. \textbf{Gap}: mean distance between a language's average score and English's, over the seven languages (the cross-lingual score shift, CSS). \textbf{Floor}: the same distance between English and reworded English. \textbf{Gap $-$ floor}: with a 95\% bootstrap interval over items; $^{\ast}$ marks intervals above zero. \textbf{Reliable drops}: languages (of seven) whose drop survives the permutation test with false-discovery correction. \textbf{CLCI} $= 1 - $ mean gap. These numbers describe consistency with English, not fairness, and should not be used to rank models for deployment.}
\label{tab:models}
\end{table*}

\begin{table*}[t]
\centering
\small
\begin{tabular}{lcccccccc}
\toprule
\textbf{Model} & \textbf{Spanish} & \textbf{French} & \textbf{Chinese} & \textbf{Arabic} & \textbf{Korean} & \textbf{Swahili} & \textbf{Hindi} & \textbf{Noise only} \\
\midrule
Aya-23-8B & 0.004 & 0.013 & 0.010 & 0.009 & 0.010 & 0.011 & 0.012 & 0.007 \\
Aya-Expanse-8B & 0.002 & 0.009 & 0.011 & 0.005 & 0.002 & 0.028 & 0.015 & 0.010 \\
BLOOMZ-7B & 0.021 & 0.019 & 0.021 & 0.010 & 0.024 & 0.005 & \textbf{0.038} & 0.010 \\
Gemma-2-9B & 0.004 & 0.017 & 0.014 & 0.010 & 0.022 & 0.012 & \textbf{0.024} & 0.009 \\
Llama-3.1-8B & 0.008 & 0.014 & 0.017 & 0.017 & 0.019 & \textbf{0.023} & \textbf{0.024} & 0.008 \\
Mistral-7B & 0.004 & 0.016 & \textbf{0.027} & \textbf{0.032} & \textbf{0.038} & \textbf{0.038} & \textbf{0.035} & 0.007 \\
OLMo-2-7B & 0.008 & 0.020 & \textbf{0.042} & \textbf{0.030} & \textbf{0.039} & \textbf{0.053} & \textbf{0.047} & 0.010 \\
Qwen-2.5-7B & 0.012 & \textbf{0.031} & 0.020 & \textbf{0.032} & \textbf{0.038} & \textbf{0.057} & \textbf{0.045} & 0.010 \\
SOLAR-10.7B & 0.002 & 0.016 & \textbf{0.027} & \textbf{0.026} & \textbf{0.034} & \textbf{0.030} & \textbf{0.028} & 0.009 \\
Yi-1.5-9B & 0.003 & 0.011 & 0.004 & \textbf{0.024} & \textbf{0.029} & \textbf{0.032} & \textbf{0.031} & 0.009 \\
\bottomrule
\end{tabular}
\caption{How far each language's average stereotype score falls below English, for every model (negative values: above English). Bold: the drop survives the permutation test with false-discovery correction. \textbf{Noise only}: the distance from zero expected if the item-level changes were pure noise, averaged over languages.}
\label{tab:gap_matrix}
\end{table*}

\begin{table}[t]
\centering
\small
\setlength{\tabcolsep}{2.4pt}
\begin{tabular}{lcccccccc}
\toprule
\textbf{Items} & \textbf{Sp} & \textbf{Fr} & \textbf{Ch} & \textbf{Ar} & \textbf{Ko} & \textbf{Sw} & \textbf{Hi} & \textbf{Avg.} \\
\midrule
All & .007 & .017 & .019 & .020 & .026 & .029 & .030 & .021 \\
Faithful & .003 & .010 & .012 & .016 & .026 & .022 & .031 & .017 \\
\bottomrule
\end{tabular}
\caption{Drop below English with all translations and with faithful ones only (back-translation similarity $\ge 0.9$, a subset of the first row). \textbf{Avg.}: equal weight per language. The reduction is 0.004 (95\% paired bootstrap interval over items [-0.002, 0.009]).}
\label{tab:controls}
\end{table}

\begin{table}[t]
\centering
\small
\setlength{\tabcolsep}{3pt}
\begin{tabular}{lcccccc}
\toprule
 & \multicolumn{2}{c}{\textbf{Drop}} & \multicolumn{2}{c}{\textbf{Reliable}} & \multicolumn{2}{c}{\textbf{Slope}} \\
\cmidrule(lr){2-3}\cmidrule(lr){4-5}\cmidrule(lr){6-7}
\textbf{Language} & avg. & sum & avg. & sum & avg. & sum \\
\midrule
Spanish & 0.007 & 0.037 & 0 & 2 & 0.37 & 0.38 \\
French & 0.017 & 0.038 & 1 & 1 & 0.38 & 0.36 \\
Chinese & 0.019 & 0.065 & 3 & 4 & 0.36 & 0.32 \\
Arabic & 0.020 & 0.057 & 5 & 5 & 0.18 & 0.24 \\
Korean & 0.026 & 0.057 & 5 & 5 & 0.14 & 0.16 \\
Swahili & 0.029 & 0.099 & 6 & 9 & 0.26 & 0.11 \\
Hindi & 0.030 & 0.069 & 8 & 6 & 0.15 & 0.21 \\
\midrule
All & 0.021 & -- & 28 & 32 & 0.26 & 0.25 \\
\bottomrule
\end{tabular}
\caption{Main results with the per-token average log-probability (avg.) and with the summed log-probability (sum), which does not dilute a one-word contrast in a long continuation. \textbf{Reliable}: models (of 10; last row: pairs of 70) with a drop that survives the corrected permutation test. Summed scores are more extreme, so drops are larger; the drop and the overall carry-over hold, while carry-over by language shifts (Swahili falls, Arabic and Hindi rise).}
\label{tab:summed}
\end{table}

\begin{table}[t]
\centering
\small
\setlength{\tabcolsep}{3.5pt}
\begin{tabular}{lcccc}
\toprule
 & \multicolumn{2}{c}{\textbf{Models agree}} & & \\
\cmidrule(lr){2-3}
\textbf{Version} & score & change & \textbf{Spread} & \textbf{Center} \\
\midrule
English & 0.50 & -- & -- & -- \\
\midrule
Spanish & 0.58 & 0.45 & 0.21 & 0.515 \\
French & 0.55 & 0.45 & 0.20 & 0.506 \\
Chinese & 0.51 & 0.38 & 0.22 & 0.503 \\
Arabic & 0.53 & 0.41 & 0.20 & 0.510 \\
Korean & 0.57 & 0.47 & 0.22 & 0.505 \\
Swahili & 0.50 & 0.50 & 0.23 & 0.496 \\
Hindi & 0.58 & 0.43 & 0.18 & 0.500 \\
\midrule
Reworded via German & 0.64 & 0.67 & 0.15 & 0.510 \\
Reworded via Finnish & 0.63 & 0.62 & 0.20 & 0.517 \\
Reworded via Japanese & 0.66 & 0.60 & 0.23 & 0.507 \\
\bottomrule
\end{tabular}
\caption{Are item-level scores outside English systematic or noise? \textbf{Models agree}: average correlation, over the 45 pairs of models, of their scores on the same items (\emph{score}) and of the change from English (\emph{change}). \textbf{Spread}: standard deviation of the item-level change. \textbf{Center}: fitted score of an item on which the model has no preference in English.}
\label{tab:agreement}
\end{table}

\begin{table*}[t]
\centering
\small
\setlength{\tabcolsep}{4pt}
\begin{tabular}{lcccccccc}
\toprule
 & & \multicolumn{2}{c}{\textbf{Reliable drops}} & \textbf{Drop} & \multicolumn{2}{c}{\textbf{Carry-over}} & \multicolumn{2}{c}{\textbf{Same choice}} \\
\cmidrule(lr){3-4}\cmidrule(lr){6-7}\cmidrule(lr){8-9}
\textbf{Items} & $n$ & soft & binary & [95\% CI] & transl. & rew. & transl. & rew. \\
\midrule
StereoSet & 200 & 41/70 & 18/70 & 0.027 [$0.016$, $0.037$] & 0.15 & 0.49 & 56\% & 67\% \\
BBQ & 125 & 0/70 & 0/70 & 0.011 [$-0.009$, $0.033$] & 0.32 & 0.72 & 61\% & 72\% \\
Written & 37 & 0/70 & 0/70 & 0.026 [$-0.019$, $0.068$] & 0.24 & 0.69 & 63\% & 82\% \\
\midrule
BBQ + Written & 162 & 0/70 & 1/70 & 0.015 [$-0.004$, $0.034$] & 0.30 & 0.71 & 61\% & 74\% \\
\bottomrule
\end{tabular}
\caption{Results by source of the items. \textbf{Reliable drops}: model--language pairs (of 70) whose drop survives the permutation test with false-discovery correction within that source. \textbf{Drop}: average drop below English with a 95\% bootstrap interval over items (the change is averaged over models and languages within each item). \textbf{Carry-over} and \textbf{Same choice}: for translations and for English rewordings. The last row excludes StereoSet entirely.}
\label{tab:sources}
\end{table*}

\begin{table*}[t]
\centering
\small
\setlength{\tabcolsep}{4.5pt}
\begin{tabular}{lccccccc}
\toprule
 & \multicolumn{2}{c}{\textbf{All items}} & \multicolumn{3}{c}{\textbf{Fully preserved only}} & \multicolumn{2}{c}{\textbf{BBQ + written}} \\
\cmidrule(lr){2-3}\cmidrule(lr){4-6}\cmidrule(lr){7-8}
\textbf{Language} & beyond rew. & $q$ & kept & beyond rew. & $q$ & beyond rew. & $q$ \\
\midrule
Spanish & .006 [$-$.009, .021] & .439 & 266/349 & .001 [$-$.017, .020] & .887 & $-$.012 [$-$.043, .020] & .635 \\
French & .016 [.000, .032] & .050 & 295/360 & .004 [$-$.013, .021] & .758 & .009 [$-$.020, .038] & .636 \\
Chinese & .018 [.001, .036] & .050 & 251/361 & .011 [$-$.008, .032] & .491 & .022 [$-$.010, .054] & .420 \\
Arabic & .018 [.003, .034] & .038 & 272/361 & .008 [$-$.009, .025] & .491 & .005 [$-$.023, .034] & .717 \\
Korean & .025 [.008, .041] & .008 & 197/361 & .022 [.002, .042] & .139 & .023 [$-$.007, .054] & .420 \\
Swahili & .028 [.010, .046] & .008 & 209/348 & .015 [$-$.006, .036] & .407 & .014 [$-$.018, .046] & .635 \\
Hindi & .029 [.015, .043] & $<$.001 & 239/357 & .028 [.010, .047] & .027 & .027 [$-$.001, .055] & .411 \\
\bottomrule
\end{tabular}
\caption{Item-level test of each language against English rewording. The change is first averaged over the ten models within each item, so that items, not model--item pairs, are the unit. \textbf{beyond rew.}: how much more the language drops below English than rewording does, with a 95\% bootstrap interval over items; $q$: paired sign-flip permutation test over items, false-discovery corrected. \textbf{Fully preserved only}: translations whose contrast was judged fully preserved (\textbf{kept}: fully preserved out of all judged translations). \textbf{BBQ + written}: all items from these two sources, without StereoSet.}
\label{tab:mixed}
\end{table*}

\begin{table}[t]
\centering
\small
\setlength{\tabcolsep}{3.5pt}
\begin{tabular}{lcccc}
\toprule
 & \multicolumn{3}{c}{\textbf{Models agree on change}} & \\
\cmidrule(lr){2-4}
\textbf{Language} & raw & $-$length & $-$length, freq. & $R^2$ \\
\midrule
Spanish & 0.45 & 0.48 & 0.47 & 0.097 \\
French & 0.45 & 0.47 & 0.46 & 0.121 \\
Chinese & 0.38 & 0.38 & 0.37 & 0.122 \\
Arabic & 0.41 & 0.41 & 0.41 & 0.067 \\
Korean & 0.47 & 0.48 & -- & 0.049 \\
Swahili & 0.50 & 0.50 & -- & 0.065 \\
Hindi & 0.43 & 0.42 & 0.42 & 0.070 \\
\midrule
\multicolumn{5}{l}{Equal length ($n$=4,778): drop 0.024, slope 0.33} \\
\multicolumn{5}{l}{All pairs: drop 0.021, slope 0.26} \\
\bottomrule
\end{tabular}
\caption{Do surface properties drive the shared item-level changes? Average correlation between models of the change from English, before and after regressing out differences in token count between the two continuations and in overall length (\emph{length}), and additionally in word frequency (\emph{freq.}; no frequency data for Korean and Swahili). $R^2$: variance in the change explained by these properties. Last two rows: items whose two continuations have the same number of tokens in both languages, and all pairs.}
\label{tab:surface}
\end{table}

\paragraph{Regression.}
An ordinary least-squares regression of the item score on resource level, model
group and item category, with standard errors clustered by item, gives
coefficients of $-0.011$ ($p = 0.019$) for mid-resource and $-0.017$
($p = 0.047$) for low-resource languages, and $-0.003$ ($p = 0.31$) and
$-0.000$ ($p = 0.86$) for the multilingual and regional model groups. Because
model group varies only across ten models, its effective sample size is ten,
not the number of items. A test across the ten models finds no difference
between families (average scores 0.518, 0.515 and 0.518; $p = 0.79$).

\section{Statistical Specification and Further Analyses}
\label{app:stats}

\paragraph{Quantities.}
For model $m$, item $i$ and condition $c$ (a language, or a rewording via German,
Finnish or Japanese), the change is $d_{mic} = \st_{mic} - \st_{mi,\text{en}}$,
defined only when the item exists in both versions and its continuations did
not collapse; every comparison uses exactly the items available in both
versions. The \emph{noise floor} of a model is
$F_m = \tfrac13 \sum_{\pi} \lvert \bar d_{m\pi} \rvert$, the mean over the three
pivots of the absolute mean change, matching the CSS, which is
$\lvert \bar d_{m\lambda} \rvert$. For the item-level test (\Cref{tab:mixed}) we
average over models, $D_{i\lambda} = \tfrac1M \sum_m d_{mi\lambda}$ and
$R_i = \tfrac1{3M}\sum_{m,\pi} d_{mi\pi}$, and test
$T_\lambda = -\overline{(D_{i\lambda} - R_i)}$ over the items present in both,
with a sign-flip permutation test (20{,}000 draws) and a bootstrap over items
(5{,}000 resamples). The sign-flip test assumes that, without a language effect,
$D_{i\lambda} - R_i$ is symmetric around zero, i.e., that the translation and the
rewording conditions of an item are exchangeable.

\paragraph{Absolute gap against the absolute floor.}
The test above concerns the signed shift. To ask whether the \emph{absolute}
gap exceeds the absolute floor, we compare $\lvert \overline{D_{i\lambda}}
\rvert$ with $\tfrac13 \sum_\pi \lvert \overline{D_{i\pi}} \rvert$ on the same
items, with a bootstrap over items. The difference is clearly above zero for
Hindi (0.023, interval 0.006--0.034) and Swahili (0.023, 0.001--0.038), just
below for Korean (0.018, $-0.003$ to 0.033), and its interval includes zero for
the other languages.

\paragraph{Injected shifts.}
For each shift size $\delta$ and each of 200 runs: (1) pick one of the 30
model--rewording pairs at random; (2) in that pair, add $\delta$ to the log-odds
$\ell(s^+) - \ell(s^-)$ of every item, which keeps scores within $(0,1)$, and
recompute its changes; (3) compute sign-flip $p$-values for all 30 pairs (2{,}000
draws) and correct them jointly; (4) record whether the shifted pair is detected
and how many of the 29 unshifted pairs are flagged. Items and scores are the
same in every run; only the shifted pair and the permutation draws vary. The
34{,}800 unshifted tests (6 sizes $\times$ 200 runs $\times$ 29 pairs) produced
no flag after correction.

\paragraph{Power.}
The required number of items is $n = (z_{0.975} + z_{0.8})^2 \sigma_d^2 /
\delta^2$, with $\sigma_d = 0.208$ the standard deviation of the item-level
change, for a single two-sided test per model--language pair; correcting over
70 pairs raises it further. For the item-level test (\Cref{tab:mixed}) we
estimate power by centring each language's item-level differences
$D_{i\lambda} - R_i$, adding a true extra drop, resampling items and rerunning
the sign-flip test (200 runs, 2{,}000 draws, $\alpha = 0.05$ before
correction): power is 0.56--0.79 across languages for 0.02 and 0.91--0.99 for
0.03.

\paragraph{Association measures.}
\Cref{tab:assoc} reports correlations, ratios of standard deviations and slopes
with bootstrap intervals over items. A least-squares slope equals the
correlation times the ratio of standard deviations. After translation the ratio
is 1.13 pooled and 0.73--1.32 by language, so translation does not compress
scores; after machine rewording it is 1.37, because round trips make some
scores more extreme. The lower slopes after translation therefore reflect lower
correlation, which holds on fully preserved translations too.

\begin{table*}[t]
\centering
\small
\setlength{\tabcolsep}{4pt}
\begin{tabular}{lccccc}
\toprule
 & \multicolumn{3}{c}{\textbf{All translated items}} & \multicolumn{2}{c}{\textbf{Fully preserved only}} \\
\cmidrule(lr){2-4}\cmidrule(lr){5-6}
\textbf{Items} & $r$ [95\% CI] & SD ratio & slope [95\% CI] & $r$ [95\% CI] & SD ratio \\
\midrule
All languages & 0.23 [0.20, 0.27] & 1.13 & 0.26 [0.22, 0.31] & 0.27 [0.23, 0.31] & 1.04 \\
\midrule
Spanish & 0.31 [0.23, 0.38] & 1.21 & 0.37 [0.28, 0.46] & 0.32 [0.23, 0.40] & 1.16 \\
French & 0.32 [0.25, 0.39] & 1.17 & 0.38 [0.29, 0.46] & 0.35 [0.27, 0.42] & 1.10 \\
Chinese & 0.28 [0.22, 0.33] & 1.32 & 0.36 [0.29, 0.43] & 0.33 [0.26, 0.38] & 1.22 \\
Arabic & 0.18 [0.12, 0.24] & 1.01 & 0.18 [0.12, 0.25] & 0.19 [0.11, 0.26] & 0.92 \\
Korean & 0.13 [0.06, 0.19] & 1.10 & 0.14 [0.07, 0.21] & 0.23 [0.15, 0.30] & 0.96 \\
Swahili & 0.20 [0.11, 0.27] & 1.27 & 0.26 [0.15, 0.35] & 0.26 [0.17, 0.35] & 1.15 \\
Hindi & 0.20 [0.14, 0.26] & 0.73 & 0.15 [0.10, 0.20] & 0.20 [0.12, 0.29] & 0.66 \\
\midrule
English rewording & 0.48 [0.43, 0.53] & 1.37 & 0.66 [0.60, 0.71] & -- & -- \\
Hindi rewording & 0.84 [0.80, 0.87] & 1.00 & 0.83 [0.78, 0.88] & -- & -- \\
Spanish rewording & 0.82 [0.76, 0.87] & 1.01 & 0.82 [0.74, 0.91] & -- & -- \\
\bottomrule
\end{tabular}
\caption{Association between an item's score before and after a change: correlation $r$, ratio of standard deviations, and least-squares slope ($= r \times$ SD ratio), each with a 95\% bootstrap interval over items. Translated rows compare English with the translation; \textbf{English rewording} compares English with its rewordings; the Hindi and Spanish rewording rows compare a translation with its rewording in the same language. Low slopes after translation come from low correlation; the SD ratio does not fall consistently below 1.}
\label{tab:assoc}
\end{table*}

\paragraph{Reversal or neutralization.}
Of model--item pairs scoring above 0.55 in English, after translation 45\%
remain above 0.55, 32\% fall between 0.45 and 0.55, and 23\% fall below 0.45
(after rewording: 68\%, 16\%, 15\%). Hindi mostly moves toward the middle (54\%),
Swahili reverses most often (31\%). Conversely, 29\% of pairs below 0.45 in
English rise above 0.55 after translation. Both reversals and moves toward
neutral contribute to the lower averages.

\paragraph{By social dimension.}
\Cref{tab:dims} pools items by the social dimension they concern. The drop is
clearest for gender (0.039, interval 0.017--0.061); for the other dimensions the
intervals include or nearly touch zero, and in every dimension translated scores
correlate with English less than reworded ones do. The StereoSet drop does not
rest on gender alone: without its gender items, StereoSet still drops 0.022
(0.010--0.033).

\begin{table}[t]
\centering
\small
\setlength{\tabcolsep}{2pt}
\begin{tabular}{lccccc}
\toprule
\textbf{Dimension} & $n$ & \textbf{Drop} [95\% CI] & $r$ tr. & $r$ rew. & \textbf{Rev.} \\
\midrule
Gender & 94 & .039 [.017, .061] & 0.20 & 0.58 & 21\% \\
Race/eth. & 82 & .013 [$-$.010, .036] & 0.27 & 0.53 & 21\% \\
Religion & 81 & .012 [$-$.008, .030] & 0.19 & 0.35 & 24\% \\
Profession & 50 & .016 [.000, .032] & 0.09 & 0.26 & 22\% \\
Age & 30 & .014 [$-$.027, .057] & 0.33 & 0.40 & 26\% \\
Nationality & 25 & .034 [$-$.002, .072] & 0.22 & 0.57 & 35\% \\
\bottomrule
\end{tabular}
\caption{Results by social dimension (pooled over sources). \textbf{Drop}: mean drop below English, bootstrap over items. $r$: correlation of item scores with English after translation and after rewording. \textbf{Rev.}: share of model--item pairs leaning stereotyped in English (score $> 0.55$) that lean counter-stereotyped after translation ($< 0.45$).}
\label{tab:dims}
\end{table}

\paragraph{BBQ ``unknown'' answers.}
In ambiguous BBQ contexts the correct answer is ``unknown''. Models rate it above
both group answers on 4\% of English items but on 6--21\% of items in other
languages (Spanish and Swahili 21\%, Korean 15\%, Hindi 13\%). Our score, which
compares only the two group answers, does not capture this shift toward the
correct answer.

\section{Reproducibility}
\label{app:repro}

\Cref{tab:flow} traces items through every stage. Models are the instruction
checkpoints \texttt{meta-llama/Llama-3.1-8B-Instruct},
\texttt{mistralai/Mistral-7B-Instruct-v0.3},
\texttt{allenai/OLMo-2-1124-7B-Instruct}, \texttt{google/gemma-2-9b-it},
\texttt{CohereForAI/aya-23-8B}, \texttt{CohereForAI/aya-expanse-8b},
\texttt{bigscience/bloomz-7b1}, \texttt{Qwen/Qwen2.5-7B-Instruct},
\texttt{01-ai/Yi-1.5-9B-Chat} and \texttt{upstage/SOLAR-10.7B-Instruct-v1.0}
from Hugging Face, loaded with 4-bit NF4 quantization (double quantization,
float16 compute). Scores use plain text with each tokenizer's default special
tokens, the per-token average log-probability of the continuation, and
continuation tokens located by character offsets. Translations and round trips
used Google Translate (April 2026). The English LLM rewordings and the contrast
judgments were produced with Claude Opus 5.5; the Hindi and Spanish rewordings
with Aya-Expanse-32B (bfloat16, greedy decoding, up to 400 new tokens, one retry
with temperature 0.7 and top-$p$ 0.95). Permutation tests use 20{,}000 draws,
bootstraps 1{,}000--5{,}000 resamples over items, with fixed seeds. Scoring ran
on one NVIDIA A100 80GB GPU (about 4 hours for all scoring and controls, and 2
hours for the within-language rewording). Code, prompts, rewordings and
judgments are released with the paper.

\begin{table}[t]
\centering
\small
\setlength{\tabcolsep}{3pt}
\begin{tabular}{p{5.6cm}r}
\toprule
\textbf{Stage} & \textbf{Count} \\
\midrule
StereoSet intrasentence items sampled & 200 \\
BBQ ambiguous items sampled / kept (roles recovered) & 150 / 125 \\
Items written by the authors & 37 \\
English items & 362 \\
Translated items (7 languages) & 2,534 \\
\quad removed: contrast collapsed & 37 \\
Translated--English pairs per model & 2,497 \\
Machine rewordings of English (3 pivots) & 1,086 \\
\quad removed: contrast collapsed & 20 \\
LLM rewordings of English sampled / usable & 100 / 92 \\
Hindi/Spanish rewordings attempted / passed & 774 / 247 \\
\quad usable after removing collapsed translations & 129 / 101 \\
Translations judged for contrast preservation & 2,497 \\
\bottomrule
\end{tabular}
\caption{Items at each stage. Every translated or reworded item is compared only with its own English original, scored by the same model.}
\label{tab:flow}
\end{table}

\section{Rewording Within a Language}
\label{app:target_rw}

Aya-Expanse-32B was asked to reword each translated Hindi and Spanish item
(context and all three continuations) in the same language, keeping the words
that differ between the stereotyped and counter-stereotyped continuation and
returning JSON. A rewording was accepted only if the two continuations still
differed, every word that distinguished them originally still appeared in the
corresponding reworded continuation, at least one field changed, and each field
kept a multilingual embedding similarity of at least 0.7 to the original; a
failed rewording was retried once with sampling. Of 774 items, 247 passed
(32\%), almost all failures coming from the strict word check, which also
rejects legitimate rewordings of items whose continuations differ in more than
one word; after removing items whose translation had collapsed, 129 Hindi and
101 Spanish items remain. The accepted items are mostly BBQ (90 Hindi, 70
Spanish) and StereoSet (36, 30). Within each source, rewording keeps more of the
preference than translation from English does (Hindi BBQ 0.86 vs.\ 0.16,
StereoSet 0.63 vs.\ 0.20; Spanish BBQ 0.83 vs.\ 0.51, StereoSet 0.63 vs.\
0.18). The accepted Hindi items are representative in their cross-language
carry-over (0.16, against 0.15 for all Hindi items); the Spanish ones carry over
somewhat better than average (0.46 vs.\ 0.37).

\begin{table}[t]
\centering
\small
\setlength{\tabcolsep}{1.6pt}
\begin{tabular}{lccc}
\toprule
\textbf{Change} & \textbf{Slope} & \textbf{Same} & \textbf{Items} \\
\midrule
\multicolumn{4}{l}{\emph{Hindi items (129)}} \\
\quad Hindi $\rightarrow$ reworded Hindi & 0.83 & 83\% & 129 \\
\quad English $\rightarrow$ Hindi & 0.16 & -- & 129 \\
\quad English $\rightarrow$ reworded English & 0.69 & -- & 129 \\
\multicolumn{4}{l}{\emph{Spanish items (101)}} \\
\quad Spanish $\rightarrow$ reworded Spanish & 0.82 & 81\% & 101 \\
\quad English $\rightarrow$ Spanish & 0.46 & -- & 101 \\
\quad English $\rightarrow$ reworded English & 0.75 & -- & 101 \\
\midrule
\multicolumn{4}{l}{\emph{StereoSet scored without the blank context}} \\
Reliable drops & \multicolumn{3}{l}{37/70 (template: 41/70)} \\
Carry-over slope & \multicolumn{3}{l}{0.18 (template: 0.15)} \\
\bottomrule
\end{tabular}
\caption{Top: carry-over of item-level preferences within a language (each translated item reworded in the same language by Aya-Expanse-32B, which is not among the audited models; 129 Hindi and 101 Spanish items passed validation), across languages, and within English by machine round-trip rewording, all on the same items. \textbf{Same}: same choice as before the change. Bottom: StereoSet items scored as sentences on their own, without the template that contains the blank.}
\label{tab:within_language}
\end{table}

\section{Pipeline Checks}
\label{app:pitfalls}

\begin{table}[t]
\centering
\small
\setlength{\tabcolsep}{2.6pt}
\begin{tabular}{lcccccc}
\toprule
 & \multicolumn{2}{c}{\textbf{English}} & \textbf{Below} & & \multicolumn{2}{c}{\textbf{Slope}} \\
\cmidrule(lr){2-3}\cmidrule(lr){6-7}
\textbf{Errors present} & soft & bin. & \textbf{En.} & \textbf{Rel.} & rew. & trans. \\
\midrule
All three errors & 0.493 & 0.42 & 35/70 & 0/70 & 0.47 & 0.21 \\
Only label errors & 0.490 & 0.44 & 33/70 & 0/70 & 0.56 & 0.28 \\
Only scorer error & 0.534 & 0.59 & 70/70 & 28/70 & 0.48 & 0.21 \\
None (ours) & 0.536 & 0.63 & 70/70 & 28/70 & 0.66 & 0.26 \\
\bottomrule
\end{tabular}
\caption{The same analysis under each combination of errors. \textbf{English}: mean soft and binary score on English items. \textbf{Below En.}: model--language pairs scoring below English; \textbf{Rel.}: of these, reliable after correction. \textbf{Slope}: carry-over for rewordings and translations. With the label errors, every gap looks like noise.}
\label{tab:pipeline}
\end{table}

\paragraph{Inverted StereoSet labels.}
In the Hugging Face release of StereoSet, label 0 marks the \emph{anti}-stereotype.
Reading it as ``stereotype'' swaps every item. \emph{Check:} read a few items
with their labels.

\paragraph{Arbitrary BBQ answer roles.}
Which BBQ option is stereotyped depends on the targeted group and on whether the
question is negative (``Who was irresponsible?'') or not; options are listed in
random order, and unknown answers are phrased in many ways (``Can't answer,''
``Not enough info''). Taking options in listed order gives each item a random
direction. Rebuilding roles from the metadata changed 61 of 125 items. The 25
others could not be rescued: an unrecognized ``unknown'' answer had taken the
place of a group answer, which was therefore never scored. \emph{Check:} derive roles from the metadata, never
from answer order.

\paragraph{The first word is not scored.}
Locating the continuation by tokenizing ``context + space'' and counting tokens
is off by one for nine of our ten tokenizers, because the space joins the next
word in the full text. The first continuation token, for ``He/She was
exhausted\ldots'' the only one that differs, is never scored. \emph{Check:}
decode the scored tokens; we locate them by character offsets instead.
Translation can also erase the contrast itself (\S\ref{sec:setup}); compare the
two continuations after translation.

\section{One Item Across Languages}
\label{app:example}

\begin{table}[t]
\centering
\small
\setlength{\tabcolsep}{2.4pt}
\begin{tabular}{lcccccccc}
\toprule
\textbf{Model} & \textbf{En} & \textbf{Es} & \textbf{Fr} & \textbf{Zh} & \textbf{Ar} & \textbf{Ko} & \textbf{Hi} & \textbf{Sw} \\
\midrule
Qwen-2.5-7B & \textbf{0.93} & \textbf{0.72} & 0.47 & \textbf{0.73} & \textbf{0.74} & \textbf{0.52} & 0.46 & 0.41 \\
Aya-23-8B & \textbf{0.83} & 0.32 & 0.42 & \textbf{0.88} & \textbf{0.78} & \textbf{0.70} & 0.43 & \textbf{0.54} \\
Mistral-7B & \textbf{0.83} & \textbf{0.51} & 0.50 & \textbf{0.77} & 0.42 & \textbf{0.54} & 0.47 & 0.33 \\
\bottomrule
\end{tabular}
\caption{One item in eight languages: \emph{``The elderly woman tried to set up her new smartphone.''} followed by \emph{``She was confused by the interface and needed constant help.''} (stereotyped) or \emph{``She configured it independently using the online guide.''} Bold: the model prefers the stereotyped continuation. Every model leans strongly toward it in English; after translation the preference holds in some languages, reverses in others, and differs between models.}
\label{tab:example}
\end{table}

\Cref{tab:example} follows one item through all eight languages. In English,
all three models strongly prefer the stereotype that an elderly woman needs
help with a smartphone (0.83--0.93). After translation the preferences remain
strong but scatter: Aya-23 keeps the stereotype in Chinese (0.88) yet reverses
it in Spanish (0.32), while Qwen keeps it in Spanish (0.72) and reverses it in
Swahili (0.41). Averaged over many such items, these choices partly cancel.

\section{Items Written in the Target Languages}
\label{app:layerb}

We also wrote 21 items directly in the seven target languages (three each), for
example about caste and regional origin in Hindi, labor migration in Arabic,
university prestige in Korean and tribal affiliation in Swahili, with an LLM
used to correct grammar. Only the Hindi items were checked by a native speaker.
They have no English counterpart, so they enter no comparison; we list them only
to illustrate what translated items miss.

\end{document}